# Cross-View kernel transfer

Riikka Huusari[a,b,*], Cécile Capponi[b], Paul Villoutreix[b,c], Hachem Kadri[b]

[a] Helsinki Institute for Information Technology HIIT, Department of Computer Science, Aalto University, Espoo, Finland
[b] Aix-Marseille University, UniversitÉ de Toulon, LIS, CNRS, Marseille, France
[c] Turing Center for Living Systems (CENTURI), Marseille, France

A B S T R A C T

We consider the kernel completion problem with the presence of multiple views in the data. In this context the data samples can be fully missing in some views, creating missing columns and rows to the kernel matrices that are calculated individually for each view. We propose to solve the problem of completing the kernel matrices with Cross-View Kernel Transfer (CVKT) procedure, in which the features of the other views are transformed to represent the view under consideration. The transformations are learned with kernel alignment to the known part of the kernel matrix, allowing for finding generalizable structures in the kernel matrix under completion. Its missing values can then be predicted with the data available in other views. We illustrate the benefits of our approach with simulated data, multivariate digits dataset and multi-view dataset on gesture classification, as well as with real biological datasets from studies of pattern formation in early *Drosophila melanogaster* embryogenesis.

## 1. Introduction

Multi-view learning is a machine learning paradigm referring to a learning situation where data contains various, often heterogenous, modalities that might be obtained from different sources or by different measurement techniques [1]. For example a dataset might contain images with captions, both of them describing the same data samples but from different points of view. Learning by taking into account all the views and their interactions is expected to give better results than learning from each single view independently, as the views are likely to carry complementary information and regularities.

Gathering multi-view data can be very expensive and in some situations (such as some biological applications, or medical diagnosis from several physical examination devices) it might be outright impossible to simultaneously measure all the views under investigation. A typical example of the latter situation arises in developmental biology when several variables are of interest but cannot be measured simultaneously [2], or when results of heterogeneous types of experiments, such as spatial information and single cell transcriptomics, need to be integrated in a common representation [3]. While many multi-view learning approaches have been developed to work directly with missing data elements for tasks such as classification or multi-view clustering among others [4–9], unfortunately many successful multi-view methods cannot directly cope with data missing from the views. The simplest approach in this case would be to neglect the samples with missing views, but depending on the amount of these samples this might make the data set so small as to make applying many of these machine learning methods non-feasible. Thus a preprocessing step to fill in the missing values is needed.

Kernel methods in multi-view learning are widely used in many fields such as computational biology and computer vision [10,11]. One especially successful and widely applied set of methods is called Multiple Kernel Learning (MKL) [12]. In kernel methods, the data samples are not considered as is by the learning algorithm, but rather via a kernel function that takes two samples and acts as a kind of similarity measure between them. This can be an especially advantageous property for the learning algorithm, as kernel functions can be defined for many types of data. For example, graphs can be difficult for many machine learning algorithms to handle, but kernel-based methods are able to treat them with no more difficulty than any other data, as the kernel-based algorithms consider the kernel matrix calculated with the samples, not the samples themselves. There are several possibilities on how to

* Corresponding author at: Helsinki Institute for Information Technology HIIT, Department of Computer Science, Aalto University, Espoo, Finland.

*E-mail addresses:* riikka.huusari@aalto.fi (R. Huusari), cecile.capponi@lis-lab.fr (C. Capponi), paul.villoutreix@lis-lab.fr (P. Villoutreix), hachem.kadri@lis-lab.fr (H. Kadri).

define kernels for many traditionally difficult data types, such as strings [13], histograms [14] or graphs [15], among others [16,17]. Thus, in this framework it is natural to directly complete the kernels themselves instead of the original missing features. Kernel completion in multi-view setting is an emerging topic which has not been much investigated so far [18].

Existing matrix completion methods can be applied to a kernel completion problem only when some individual kernel values are missing, and not the whole rows and columns. More often than not, in our setting the missing values span indeed whole rows and columns, and regular matrix completion approaches cannot cope with the completion task. In order to succeed in filling in the values, the multi-view structure of the data should be leveraged for kernel completion. In this paper we propose a novel method for problem of multi-view kernel completion, that is based on the idea of information transfer across the views. One assumption in multi-view learning is that there are some relationships between the views; the views are connected and they describe the same data, they are not fully independent. In our method we learn and transfer the information that other views contain to represent the view we wish to complete. We consider the features of the other views and align their transformation to known values we have in the kernel of the view we wish to complete, using the notion of kernel alignment [19,20]. When we have learned this transformation, we can predict the missing values based on the information in the other views. Our method is a very general in the sense that we do not require any of the views to be complete; all of them may have some missing data.

Going beyond this assumption, [21] and [22] have proposed methods filling in missing values of multi-view kernel matrices. Both of these methods hinge crucially on treating the kernel matrices as combinations of each other, something we do not consider in our approach.

Cross-view learning, or learning mappings between the views, has been previously considered in the deep learning regime for missing view imputation in [23,24]. Of these, [24] considers adversarial encoder-decoder architecture, and [23] convolutional neural networks to work with image data. These works operate in a very different regime than our proposition – as deep learning methods, they require large amounts of data, and are restricted in the types of data they can accept as input. Moreover, [24] considers only two views, while our method generalizes to any number of views.

Previous work has shown that it is possible to use a linear transformation on the kernel matrix to learn optimal domain adaptation [25]. This transformation is similar to ours, however in [25] the features to be transformed were explicitly fixed to be empirical features obtained from kernel matrix, and instead of optimizing with respect to kernel alignment they considered Hilbert-Schmidt independence criterion [26]. In contrast to our work, the idea of transforming the features was considered in the context of domain adaptation, where the goal was to learn a common feature representation given kernel containing data from two domains. In our case the transfer is done from multiple feature representations to one that describes still another kernel.

This paper is organized as follows. The next section introduces relevant background about related works and kernel methods. Section 3 introduces our algorithm (called CVKT for Cross-View Kernel Transfer), which we validate with experiments on simulated and real data in Section 4. Our experiments have a two-fold focus: first of all to show the validity of our method from the point of view of kernel completion, and secondly we aim to show the applicability also when the completed kernels are consequently used in classification. For the first goal, of particular interest to us is a set of real biological data from studies of pattern formation in early *Drosophila melanogaster* embryogenesis, in part motivating our work. Section 5 concludes and discusses possibilities for future work.

## 2. Background

We now discuss more in depth the problems of matrix and kernel completion, in both traditional and multi-view settings. We then follow with short introduction to kernel methods.

We denote scalars, vectors and matrices as $a$, $\mathbf{a}$ and $\mathbf{A}$, respectively. We consider the sample size to be $n$, and denote number of views in the data with $V$. The view of e.g. a matrix is indicated in parenthesis in superscript, as $\mathbf{M}^{(v)}$. We denote $\langle\cdot,\cdot\rangle_F$ and $||\cdot||_F$ the Frobenius inner product and norm over matrices, and $\mathbf{M}^\top$ denotes the matrix transpose.

### 2.1. Multi-view kernel matrix completion

Dealing with missing samples or features is a much studied problem in data sciences. Missing data often refers to missing feature values in the dataset, for example in a recommendation system a feature of a data sample is missing if an user has not given a rating to one item in the catalogue. Usually the data samples are stacked in a matrix, and the matrix structure is used in filling in the missing values here and there in the matrix. Matrix completion approaches often consider a low-rank approximation with which the missing values are inputed [27,28]. In addition to matrices built directly from the features, matrix completion can be used in filling in individual missing values in a kernel matrix. However matrix completion is not always applicable to kernel completion, since kernel matrices have properties (symmetry, positiveness of eigenvalues) that matrix completion algorithms might not guarantee to preserve.

Matrix completion usually deals with only one set of data, and thus there are some restrictions in the ways the data can be completed. For example every data sample must contain some features, and every feature must be present in some samples. In other words, there cannot be fully missing data samples or features, or fully missing rows or columns in the matrix. Of course in most settings if a data sample is fully missing no algorithm can recover it. However if there is some additional information available, even this can be done. Data completion in multi-view setting uses the complementary information from the views as this sort of additional information. Even here, filling in a fully missing data sample completely is a challenge. As kernel methods are prominent in multi-view learning, the kernel matrices containing similarities between data samples can be filled instead, giving rise to multi-view kernel matrix completion. It is reasonable to predict the similarities in a view where some of them are missing based on the information available in the other views – as the various views are related to each other, so are the subsequent kernel matrices. The standard assumption in the multi-view learning paradigm is that the views are correlated with each other. Even if our method does not explicitly use the mathematical formulation of correlation, it heavily relies on the relationships between the views in the data in order to build the linear feature transformations across the views.

First works for completing kernels of multiple views contain relatively restrictive assumptions, requiring one complete observed view [29,30]. Going beyond this assumption, [21] proposed an EM-algorithm that minimizes the KL-divergence of all the individual view matrices to their linear combination. Lastly, a framework for completing kernel matrices in multi-view setting has been proposed in [22], where both within- and between-view relationships are considered in solving the problem. As within-view relationship they learn a low-rank approximation of the kernel based on the available values there, while the between-view relationship strat-egy is based on finding a set of related kernels for each missing entry and modelling the kernel as a weighted sum of those matrices. In contrast to these works, our method directly considers the data interactions in the other views, and predicts the missing data in a kernel matrix with them. The work of [8] considers multi-view learning with kernels and in their framework presents a way to deal with missing data. However the completion they are interested in is done in a specific landmark space, and not on the kernel values we wish to complete.

Some works use matrix completion methods in multi-view setting in predicting the labels of a supervised learning problem [31,32]. These approaches stack the multi-view data with their labels in a large matrix, and complete the test data labels. Usually this is done for multi-output predictions, and this transductive learning setting (only the labels are learned) is very distinct from our problem; we consider unsupervised setting where kernel values on the data are learned without considering the associated labels.

It is also possible to bypass the problem of matrix completion completely, if one uses learning methods that are able to take into account the missing views. For example for incomplete multi-view clustering, methods learning a latent space via e.g. matrix factorization [4], consensus graph [6] or with generative adversarial networks [5]. In supervised setting, works adapting to incomplete multi-view data include for example a landmark-based SVM method [8], deep networks [9] and in the context of weakly labeled multi-label data [7].

### 2.2. Learning with kernels

We introduce here relevant background of kernel methods, and the notation we use in this paper in developing our method to solve the kernel completion problem. We consider multi-view data $x \in \mathcal{X} = \mathcal{X}^{(1)} \times \ldots \times \mathcal{X}^{(V)}$ such that each (complete) data sample $x$ is observed in $V$ views, $x = (x^{(1)}, \ldots, x^{(V)})$.

In machine learning kernel methods are a very successful group of methods used in various tasks [16]. The main advantage of using a kernel function $k : \mathcal{X} \times \mathcal{X} \to \mathbb{R}$ in a learning task comes from the fact that it corresponds to an inner product in some feature space (more concretely in the reproducing kernel Hilbert space (RKHS) $\mathcal{H}$ induced by the kernel), that is,

$$k(x, z) = \langle \phi(x), \phi(z) \rangle_{\mathcal{H}}.$$

This allows one to map data inexpensively to some (possibly infinite-dimensional) feature space where the data is expected to be better represented. In kernel-based learning algorithms the data is always dealt with via the kernel function so this feature representation is never explicitly needed. In practice a matrix, $\mathbf{K}$, is built with the kernel function applied to all pairs of data samples such that $\mathbf{K}_{ij} = k(x_i, x_j)$.

For multi-view learning the simplest and most widely used kernel-based approach is to build the kernel as a combination of kernels from individual views. This combination is usually a weighted sum

$$k(x, z) = \sum_{v=1}^{V} \alpha^{(v)} k^{(v)}\left(x^{(v)}, z^{(v)}\right), \tag{1}$$

where the weights $\alpha^{(v)}$ are often learned (multiple kernel learning, MKL) [12]. Whenever there is some missing data in the views, obviously the sum cannot be calculated and the corresponding values in the final kernel matrix will be missing, too. This is illustrated below, where grey areas of the kernel matrices indicate that the values are available, and white areas thus unknown.

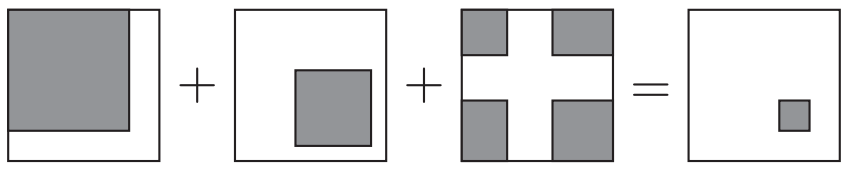

The goal of our work is to fill in these missing values in the kernel matrices by using the multi-view properties of the data, and leveraging the information contained in the other views in completing the missing values of a view.

Our kernel completion method is based on the idea of trying to form a kernel matrix as similar as possible to the one under completion by transforming features from other views. In order to do this, we need a way to compare two kernel matrices. We choose to use the notion of kernel alignment [19,20] as the similarity measure between two kernel matrices. Alignment between two matrices $\mathbf{M}$ and $\mathbf{N}$ is defined as

$$A(\mathbf{M}, \mathbf{N}) = \frac{\langle \mathbf{M}_c, \mathbf{N}_c \rangle_F}{\|\mathbf{M}_c\|_F \|\mathbf{N}_c\|_F}, \tag{2}$$

where subscript $c$ refers to centered matrices, that is, $\mathbf{M}_c = \mathbf{CMC}$ where $\mathbf{C} = \left[\mathbf{I}_n - \frac{1}{n}\mathbb{1}_n\mathbb{1}_n^\top\right]$ with $\mathbf{I}_n$ the identity matrix, $\mathbb{1}_n$ vector of ones, and $\mathbf{M}$ is of size $n \times n$. Kernel alignment has been successfully used in kernel learning problems for classification and regression, when kernel alignment has been used to match the kernel to be learned with a so-called ideal kernel calculated with labels of the learning task ($\mathbf{y}\mathbf{y}^\top$). This approach is expected to produce good predictors [20].

## 3. Cross-View kernel transfer algorithm

We propose to fill in the missing values in multi-view kernel matrices by transferring the information available in other views to represent the view in question. Contrary to other approaches based on *treating/processing* the view interactions as linear combinations of the kernels on views (or some quantity tied to the kernels), ours directly considers the features and feature interactions, and based on those is able to predict the missing views.

### 3.1. Building blocks of cross-view transfer

Given a multi-view data set $X^{(1)}, \ldots, X^{(V)}$ containing $n$ samples, we can build a $n \times n$ kernel matrix for each of the views, $\mathbf{K}^{(1)}, \ldots, \mathbf{K}^{(V)}$. Kernel-based learning algorithms take these kernels instead of original data samples when solving the learning problem.

As mentioned, a kernel corresponds to an inner product of data samples mapped to some feature space. If we know the feature map the kernel uses, we can stack the features $\phi(x_i)$, into a matrix $\Phi^{(v)}$ of size $n \times f$, with $f$ the dimensionality of the feature space. We can then write $\mathbf{K}^{(v)} = \Phi^{(v)}[\Phi^{(v)}]^\top$. For example with linear kernel we would have $\Phi^{(v)} = \mathbf{X}^{(v)}$ and $\mathbf{K}^{(v)} = \mathbf{X}^{(v)}[\mathbf{X}^{(v)}]^\top$. Of course if the feature map is infinite-dimensional (as is the case with Gaussian kernel, for example), it is not possible to stack the data projections into a matrix. However the $\Phi^{(v)}$ is not unique, and for a set of samples it is usually easy to find an alternative feature map producing the same kernel matrix. For any kernel matrix, the empirical feature map [33] defined as $\hat{\Phi}^{(v)} = \mathbf{K}^{(v)}(\mathbf{K}^{(v)})^{-1/2}$ is equally valid choice that produces the same kernel matrix, since

$$\hat{\Phi}^{(v)}[\hat{\Phi}^{(v)}]^\top = \mathbf{K}^{(v)}[\mathbf{K}^{(v)}]^{-1/2}[\mathbf{K}^{(v)}]^{-1/2}\mathbf{K}^{(v)} = \mathbf{K}^{(v)}[\mathbf{K}^{(v)}]^{-1}\mathbf{K}^{(v)} = \mathbf{K}^{(v)}.$$

Due to the fact that the empirical feature map is easy to obtain for any kernel, our method is applicable no matter what the kernels of the views are.

It is also possible to approximate the feature map, for example through Nystrŋm approximation scheme [34] which is widely used in approximating kernel matrices. Nystrm approximation is obtained by randomly sampling $m < n$ data samples, and with those calculating $\mathbf{K}^{(v)} \approx \mathbf{K}_{:,P}^{(v)}[\mathbf{K}_{P,P}^{(v)}]^{-1}\mathbf{K}_{P,:}^{(v)}$ where subscript $P$ denotes the set of these $m$ samples. In this case $\widetilde{\Phi}^{(v)} = \mathbf{K}_{:,P}^{(v)}[\mathbf{K}_{P,P}^{(v)}]^{-1/2}$ and $\mathbf{K}^{(v)} \approx \widetilde{\Phi}^{(v)}[\widetilde{\Phi}^{(v)}]^\top$. This is the approach we follow in our experimental section, however any proper kernel approximation is equally valid to be used in our algorithm.

**Table 1**
The components in the CVKT model, their sizes/lengths and corresponding explanations. Here $m^{(v)}$ the dimensionality of the (possibly approximated) features of the $v$th view, and $r$ is the rank (or number of columns) chosen for the transformation matrix.

| Notation | Size | Explanation |
|---|---|---|
| $\mathbf{K}^{(v)}$ | $n \times n$ | Kernel matrix on view $v$ |
| $\Phi^{(v)}$ | $n \times m^{(v)}$ | Matrix of features on view $v$, with $\Phi^{(v)}[\Phi^{(v)}]^\top = \mathbf{K}^{(v)}$ |
| $\mathcal{I}^{(v)}$ | $i^{(v)}$ | Set of observed samples of view $v$ |
| $\mathbf{K}_{\mathcal{I}}^{(v)}$ | $i^{(v)} \times i^{(v)}$ | Kernel matrix on the observed samples of view $v$ |
| $\Psi^{(v)}$ | $n \times \left[\sum_{j=1}^{V} m^{(j)} - m^{(v)}\right]$ | Matrix of concatenated feature representations from all views but $v$, on all samples |
| $\Psi_{\mathcal{I}}^{(v)}$ | $i^{(v)} \times \left[\sum_{j=1}^{V} m^{(j)} - m^{(v)}\right]$ | Matrix of concatenated feature representations from all views but $v$, on the samples known in view $v$; $\mathcal{I}^{(v)}$ |
| $\mathbf{U}^{(v)}$ | $\left[\sum_{j=1}^{V} m^{(j)} - m^{(v)}\right] \times r$ | Matrix transforming the features in $\Psi_{\mathcal{I}}^{(v)}$ |

Obviously, the kernel matrix $\mathbf{K}^{(v)}$ contains missing rows and columns if some of the data is missing for this view. We denote the set of indices where data is available for view $v$ as $\mathcal{I}^{(v)}$, and the size of the set as $i^{(v)} \leq n$. Whenever clear from the context which view is in question we might leave the superscript out, denoting $\mathcal{I}^{(v)} = \mathcal{I}$. We denote the section of the kernel matrix of view $v$ containing the known values as $\mathbf{K}_{\mathcal{I}}^{(v)}$; this is a matrix of size $i^{(v)} \times i^{(v)}$. We have summarized this notation (among the notation introduced in next section describing the CVKT algorithm) in Table 1.

### 3.2. Cross-View kernel transfer algorithm

We propose to learn to represent the kernel $\mathbf{K}_{\mathcal{I}}^{(v)}$ with the features of other views, and their interactions. We can leverage the kernel matrices available in other views and obtain the (empirical) features for the data samples, which we use for predicting the missing values of $\mathbf{K}^{(v)}$. To transfer the knowledge from other views towards the view $v$ under question, we firstly build a large feature matrix from the feature matrices of all the other views as

$$\Psi_{\mathcal{I}}^{(v)} = \left[\Phi_{\mathcal{I}^{(v)}}^{(1)}, \dots, \Phi_{\mathcal{I}^{(v)}}^{(v-1)}, \Phi_{\mathcal{I}^{(v)}}^{(v+1)}, \dots, \Phi_{\mathcal{I}^{(v)}}^{(V)}\right]. \tag{3}$$

Note that the features of the view under completion task are naturally left out from this matrix. From each view we take to this matrix only the samples that are available in view under study, $\mathcal{I}^{(v)}$. The new feature matrix $\Psi_{\mathcal{I}}^{(v)}$ is thus of size $i^{(v)} \times \left(m^{(1)} + \ldots + m^{(v-1)} + m^{(v+1)} + \ldots + m^{(v)}\right)$. If features are missing from some views in $\Psi_{\mathcal{I}}^{(v)}$, they are inputed there with zeros to indicate this. While we do not assume that no data sample has to have complete view in general, we do assume that at least one other view is observed at the same time with view under completion. Thus every row in $\Psi_{\mathcal{I}}^{(v)}$ contains at least some features from other views, even if some are missing. This procedure is illustrated in Fig. 1.

Learning to represent the target kernel $\mathbf{K}_{\mathcal{I}}^{(v)}$ with $\Psi_{\mathcal{I}}^{(v)}$ is done by considering a linear transformation of these features to some other feature space. This transformation is defined by matrix $\mathbf{U}^{(v)}$ of size $\left(m^{(1)} + \ldots + m^{(v-1)} + m^{(v+1)} + \ldots + m^{(V)}\right) \times r$. Here $r$ refers to "rank" of the transformation and should be less than, or equal to $m^{(1)} + \ldots + m^{(v-1)} + m^{(v+1)} + \ldots + m^{(V)}$, and is chosen when the CVKT algorithm is called. In essence, the parameter $r$ tells what is the dimension of the transformed features representing the kernel $\mathbf{K}^{(v)}$. We wish to learn the optimal transformation $\mathbf{U}^{(v)}$ such that the transfer kernel $\Psi_{\mathcal{I}}^{(v)}\mathbf{U}^{(v)}[\Psi_{\mathcal{I}}^{(v)}\mathbf{U}^{(v)}]^\top$ is maximally aligned to the target kernel, giving us the optimization problem

$$\max_{\mathbf{U}^{(v)} \in S} \quad A\left(\mathbf{K}_{\mathcal{I}}^{(v)}, \Psi_{\mathcal{I}}^{(v)}\mathbf{U}^{(v)}\left[\Psi_{\mathcal{I}}^{(v)}\mathbf{U}^{(v)}\right]^\top\right), \tag{4}$$

where we regularize the transformation matrix $\mathbf{U}^{(v)}$ by constraining it to the sphere manifold $S$, meaning that $\|\mathbf{U}^{(v)}\|_F = 1$. The optimization problem can be solved with gradient-based approach. We implemented this with the Pymanopt package [35].[1]

We wish to highlight the fact that our transformation is very general, and indeed much more powerful than simply re-weighting the views. Our approach learns, in a sense, one transformation for each view other than $v$. Yet these transformations are learned jointly in $\mathbf{U}^{(v)}$, ensuring the overall quality of the alignment. This also means that our method is capable of learning if one view should be favoured over the others, for example, or more general relationships between the views.

After solving this optimization problem, a prediction on the full kernel matrix can be done via selecting all the other views to $\Psi^{(v)}$ as

$$\Psi^{(v)} = \left[\Phi^{(1)}, \dots, \Phi^{(v-1)}, \Phi^{(v+1)}, \dots, \Phi^{(V)}\right] \tag{5}$$

and calculating

$$\tilde{\mathbf{K}}^{(v)} = \Psi^{(v)}\mathbf{U}^{(v)}\left[\Psi^{(v)}\mathbf{U}^{(v)}\right]^\top. \tag{6}$$

We summarize the Cross-view Kernel Transfer (CVKT) procedure in Algorithm 1.

**Algorithm 1** CVKT algorithm

**Require:** Set of kernels $\mathbf{K}^{(1)}, \ldots, \mathbf{K}^{(V)}$; indices of known values $\mathcal{I}^{(1)}, \ldots, \mathcal{I}^{(V)}$; parameter $r$ to control the size of transformation matrices $\mathbf{U}^{(v)}$
**for** $v \in [1, \ldots, V]$ **do**
  Calculate feature representation $\Phi_{\mathcal{I}}^{(v)}$ from $\mathbf{K}_{\mathcal{I}}^{(v)}$
**end for**
**for** $v \in [1, \ldots, V]$ **do**
  Build $\Psi_{\mathcal{I}}^{(v)}$ and $\Psi^{(v)}$ as in Eqs. 3 and 5
  Solve for $\mathbf{U}^{(v)}$ in Eq. 4
  Predict $\tilde{\mathbf{K}}^{(v)}$ with $\Psi^{(v)}$ and $\mathbf{U}^{(v)}$ as in Eq. 6
**end for**
**return** $\tilde{\mathbf{K}}^{(1)}, \ldots, \tilde{\mathbf{K}}^{(V)}$

It is important to note that we do not assume that the views used in completing the other are fully observed. We assume that each data sample is fully observed in at least one other view, and that each view contains some observed data samples. Thus $\Psi^{(v)}$ will always have some observations available in every row to which we can apply the transformation. In learning the transformations, we fill in the missing values in the features in $\Psi^{(v)}$ with zeros, as shown in Fig. 1. When learning the transformation matrix $\mathbf{U}^{(v)}$, the zero values in features have no effect on it; the areas of $\mathbf{U}^{(v)}$ that would be affected by this feature will be multiplied with zero, and in a sense left out in the decision process. (Note however that there is always at least one view available to learn with, as per our assumption.) Thus when learning $\mathbf{U}^{(v)}$, it also learns which view combinations work together and how. From this we can see that the structure of missing data distribution can affect the transformation, as after training CVKT expects to use only certain subsets of views in predicting kernel values. More concretely, the missing data distributions should be the same in training and testing for CVKT to be able to generalize. For example let us consider a dataset with three views, 0, 1 and 2, from which we want to fill in missing values in view 0. If view 1 only has samples available where 0 does, and view 2 only where 0 does not, CVKT naturally will not be able to learn a predictive mapping from view 2 to 0 as there are no training samples for this configuration. The same logic applies similarly also to other settings, for example if view 1 is as described above and view 2 is full, CVKT should be trained only with view 2. Otherwise in training it would learn a mapping $\{1, 2\} \to \{0\}$, while it should predict $\{2\} \to \{0\}$.

[1] The CVKT code is available at RH's personal website.

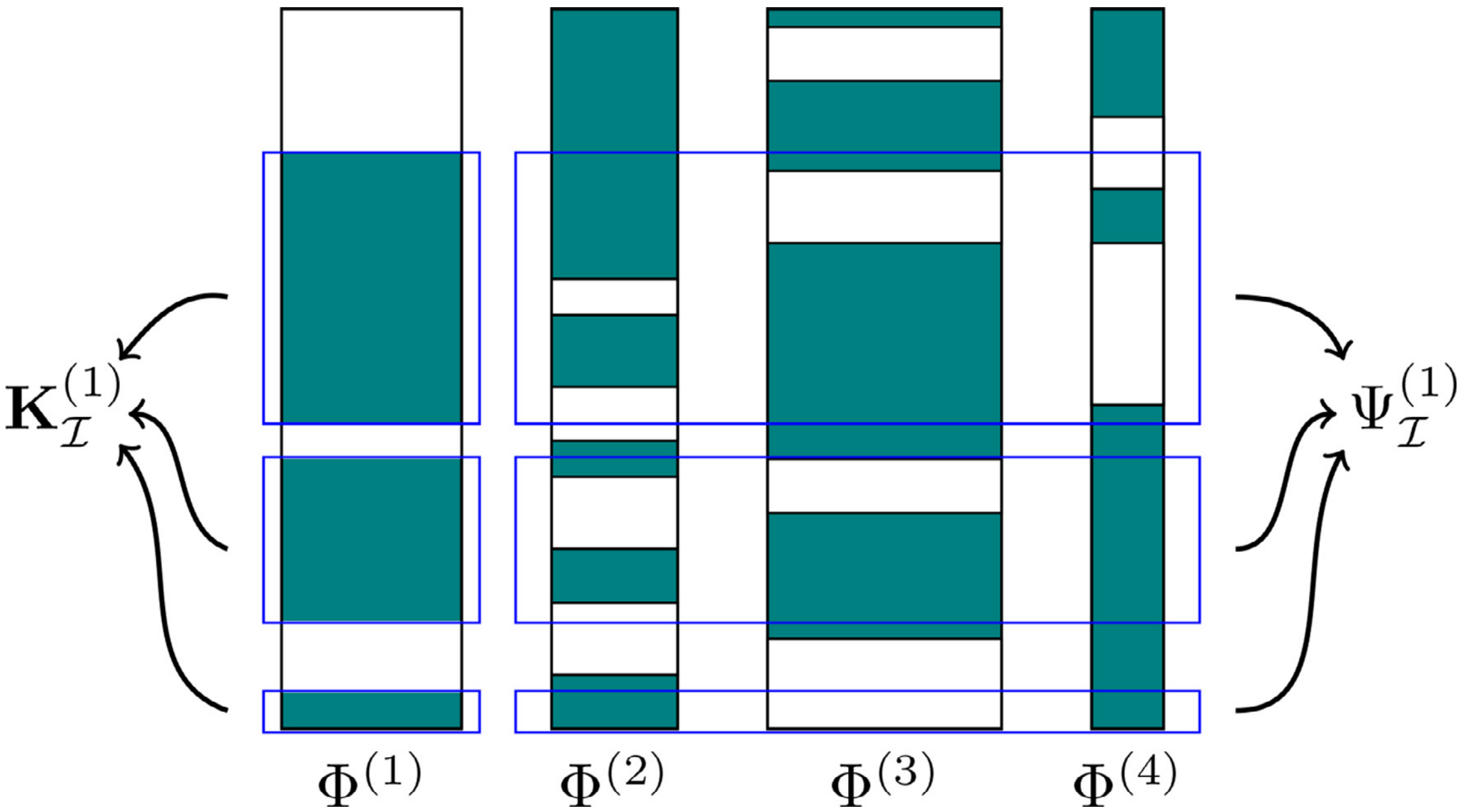

**Fig. 1.** Illustration on building the feature matrix $\Psi_{\mathcal{I}}^{(1)}$ (see Eq. (3)) in our method from the feature representations $\Phi^{(2)} - \Phi^{(4)}$. The white areas represent the missing data, and are filled with zero-inputation.

Multi-view learning paradigm focuses on data where different representations (or views) are drawn from one source. The various views describe different aspects of the same data, and may contain complementary information to each other. As the views are drawn from the same source, it is to be expected that they agree in predictive tasks (consistency). In unsupervised learning settings (such as our work for the unsupervised task of multi-view kernel completion), it can be difficult to talk about view agreement, since there is no prediction task in which the views can agree. Yet we argue that our alignment-based optimization problem promotes consistency between the views. One can see the maximal alignment between $\mathbf{K}_{\mathcal{I}}^{(v)}$ (the kernel matrix on available data, to be completed) and $\Psi_{\mathcal{I}}^{(v)}\mathbf{U}^{(v)}\left[\Psi_{\mathcal{I}}^{(v)}\mathbf{U}^{(v)}\right]^{\top}$ (the kernel matrix built from feature representations of other views) as promoting consistency between the views: the transformation learns to match the different views as well as possible.

Compared to the only two other approaches for multi-view kernel matrix completion [21,22], CVKT differs in the basic optimization procedure. The other approaches treat the optimization jointly over all the views, meaning that all the values have to be completed at once, while CVKT treats the view completion problems independently, one view at a time. Therefore CVKT can be applied to kernel completion problems more flexibly. Moreover, the other approaches only consider that the views are interacting via linear combinations over the whole views; our algorithm works in transforming a full feature space concatenated over set of views: its applicability is broader. The transformation we learn on the kernel features is very expressive, and can be expected to learn complicated relationships between the views, and thus to adapt to complementary views better than the more restrictive model of representing the kernel matrices as linear combinations of each other.

The complexity of the CVKT algorithm is naturally dependent on the number of samples available in the view processed at each iteration, $i^{(v)}$, meaning that our algorithm is faster with more missing data. The other two important parameters, $m^{(v)}$ for the feature dimensions, and $r$ for the number columns in $\mathbf{U}^{(v)}$ can be pre-set or cross-validated. As CVKT is solved with gradient-based method we consider the complexity of calculating the derivative of (4) w.r.t $\mathbf{U}^{(v)}$. The derivative is straight-forward to calculate, and the complexity arises from simple matrix multiplications. The matrix multiplications can be performed in various orders, and the preferred order depends on which variables are assumed to be small. For convenience, let us denote $m = m^{(1)} + \ldots + m^{(v-1)} + m^{(v+1)} + \ldots + m^{(V)}$. Recall that $r \leq m$. If we further assume that it is very small (i.e. $r \ll m$), and that the feature approximations are relatively small (i.e. $m < i^{(v)}$, we can calculate the gradient in $\mathcal{O}([i^{(v)}]^2 m)$.

## 4. Experiments

In this section we empirically validate our approach (CVKT) in order to illustrate and validate its properties and performance.[2] In our experiments we aim to show that CVKT performs the kernel matrix completion accurately, and we do this with simple simulated data alongside with a real dataset from study of pattern formation in *Drosophila melanogaster* embryogenesis. We further show its utility for classification problems with multi-view datasets containing also class labels (handwritten digits and time-series data on gestures). Our results show that using CVKT-inputed kernel matrices in learning problems will yield superior performance w.r.t classification accuracy, compared to other ways to fill in the data in the kernel matrices. This shows that our kernel completion results, while being accurate with respect to completion error measures, are also suitable to be used in consecutive machine learning tasks.

[2] The CVKT code and the datasets used in the experiments are available at riikkahuusari.com.

### 4.1. Compared methods

There are very few works in multi-view kernel completion setting, and very few relevant methods to compare ours to. Taking example from another paper solving multi-view kernel matrix completions problem [21], we compare our method to two simple baselines; mean and zero imputation, where the missing values are replaced with kernel mean value, or zeros, respectively. Additionally, we also consider the more elaborate MKC [22] method, and use the code provided. [3] From the methods introduced in the paper, we focus on $\text{MKC}_{emdb(ht)}$, as it is very general in the sense that it is intended to be used when kernel functions in different views are not the same and the kernel matrices have different eigenspectra. In their experiments, [22] have considered as a competing method an EM-based algorithm. However it operates with more restrictive assumptions than our algorithm, requiring a view where there are no missing samples present. In order for us to use this method, we would need to make our experimental setting considerably easier than that which our paper considers, and thus we have left it out.

Going beyond the specific area of multi-view kernel matrix completion, many methods exist that work with incomplete multi-view data. For example for classification with kernel methods, [8] adapts a landmark-based approach, and provides also an extension for adapting the method to the case with missing samples in the data. Unfortunately this method assumes that the landmarks are fully observed under all the views, which is not applicable to our experimental setting where each view can have missing samples, and each data sample can have missing views.

In the multi-view clustering literature there are many works dealing with missing views. One line of work in this context is based on nonnegative matrix factorization (NMF). While clustering with incomplete views is very different from the problem of kernel matrix completion tackled in this paper and thus comparing for completion accuracy is not possible, we can nevertheless make some comparisons to this approach. Namely, as these methods build a common representation of the views, we can use this common representation in classification task, instead of applying k-means clustering on it. Thus, we consider the MIC method presented in [4] as a competitor for our classification experiments. Even with this change to the method the settings are still very different: while with the other methods we can use the individual views completed with the different schemes, with MIC we only have the common representation from all the views.

We wish to highlight that NMF applied on the individual views is not applicable in missing views setting by itself, since in this case the whole row of data is missing. Moreover, the NMF approaches assume that the features for the data are available, which is something we do not require (we require only the incomplete kernel matrices). Also, the NMF methods require vectorial data from all the data views, while as a kernel method our CVKT can handle views of widely different data types, as long as a kernel can be defined on them.

### 4.2. Experimental protocols

In all CVKT experiments we use features extracted with Nystrm approximation, and cross-validate over different approximation levels (20%, 40%, ..., 100%). We also cross-validate over the rank (or number of columns $r$) of matrices $\mathbf{U}^{(v)}$, over similar intervals (20%, 40%, ..., 100% of the full rank $m$). For MKC, we performed the cross-validation over the parameters suggested in the code ($c_1 = [1000]$, $c_2 = [1, 10]$ and $c_3 = [0.001, 0.01, 0.1, 1, 10]$), adding values 10 and 0.1 for $c_1$. With MIC method we fix $\alpha = \beta = 0.01$ for all the views as suggested [4]. We cross-validate over the "mean fit ratio" and "error" parameters, in [0, 0.2, 0.4, 0.6, 0.8, 1] and [0.1, 0.01, 0.001, 0.0001], respectively, and use random initialisation for initial NMFs for the views. In choosing the best results in cross-validation we used the CA error measure defined below. For all our experiments we choose the samples assumed to be missing randomly, taking care that no view or sample would go fully unobserved.

For measuring the unsupervised kernel completion performance, we consider the metrics in the two other multi-view kernel matrix completion papers; the *completion accuracy (CA)* in [21] and *average relative error (ARE)* in [22]. The CA error measure is defined as

$$CA = \frac{1}{V}\sum_{v=1}^{V}\left(1 - \frac{\mathrm{Tr}\left(\mathbf{K}_{true}^{(v)}\mathbf{K}_{pred}^{(v)}\right)}{\left\|\mathbf{K}_{true}^{(v)}\right\|_F\left\|\mathbf{K}_{pred}^{(v)}\right\|_F}\right), \tag{7}$$

and the ARE over one view as

$$ARE = \frac{1}{n^{(v)} - i^{(v)}}\sum_{t\notin\mathcal{I}^{(v)}}\frac{\left\|\mathbf{K}_{pred}^{(v)}[t,:] - \mathbf{K}_{true}^{(v)}[t,:]\right\|_2}{\left\|\mathbf{K}_{true}^{(v)}[t,:]\right\|_2}, \tag{8}$$

where $\mathbf{K}_{true}^{(v)}$ and $\mathbf{K}_{pred}^{(v)}$ are the correct and the predicted kernel matrices on view $v$, respectively, and $[t,:]$ refers to the row $t$ of the kernel matrix. Unlike CA, the error measure ARE is only computed over the rows corresponding to the originally missing samples. We also consider Frobenius norm error, that is,

$$Fro = \|\mathbf{K}_{true}^{(v)} - \mathbf{K}_{pred}^{(v)}\|_F / \|\mathbf{K}_{true}^{(v)}\|_F. \tag{9}$$

Compared to ARE, this measure considers also the already known rows of kernel matrix. In all of these error measures lower value means better completion performance. In addition to these two measures, we use the structural similarity index [36], defined as

$$S.sim = \frac{\left(2\mu_{\mathbf{K}_{true}^{(v)}}\mu_{\mathbf{K}_{pred}^{(v)}} + c_1\right)\left(2\sigma_{\mathbf{K}_{true}^{(v)}\mathbf{K}_{pred}^{(v)}} + c_2\right)}{\left(\mu^2_{\mathbf{K}_{true}^{(v)}} + \mu^2_{\mathbf{K}_{pred}^{(v)}} + c_1\right)\left(\sigma^2_{\mathbf{K}_{true}^{(v)}} + \sigma^2_{\mathbf{K}_{pred}^{(v)}} + c_2\right)}, \tag{10}$$

in which $\mu_x$ is mean of $x$, $\sigma_x$ is variance of $x$, $\sigma_{xy}$ is covariance of $x$ and $y$, and $c_1$ and $c_2$ are variables for stabilizing the division (see [36]). It is a measure dedicated for image comparisons, in which properties like luminance or contrast do not affect the comparison result since they do not affect the structure of the image. For structural similarity index (*s.sim*) a high value means that the two images are similar.

In the second set of our experiments with labeled multi-view data, we use the traditional classification accuracy in assessing the performance of our method. We further validate these results with the McNemar's test of statistical significance.

Our method is expected to find generalizable structures on the kernel and predicting them in the completed matrices. It is important to notice that while this is the case, the original known values of the kernel are not necessarily fully preserved in the learned kernel. Thus in all the experiments we perform post-processing on the kernel predicted with CVKT by scaling the kernel values to the range of values in original kernel matrix, and shifting it so that the mean is the same as in the known part of the original kernel.

[3] https://github.com/aalto-ics-kepaco/MKC_software

**Table 2**
The kernel completion results on simulated data averaged over the seven views in the data with various amounts of missing views per data sample ($a$). The arrow below error measure shows whether higher values (↑), or lower values (↓) indicate superior performance.

| Error measure | a | CVKT | MKC | zero-input. | mean-input. |
|---|---|---|---|---|---|
| CA | 1 | **0.010±0.003** | 0.071 ± 0.065 | 0.143± 0.039 | 0.015± 0.006 |
| (↓) | 2 | **0.012±0.002** | 0.054± 0.024 | 0.285± 0.048 | 0.027± 0.007 |
| | 3 | **0.015±0.004** | 0.114± 0.058 | 0.427± 0.049 | 0.038± 0.009 |
| | 4 | **0.025±0.002** | 0.309± 0.062 | 0.571± 0.043 | 0.047± 0.011 |
| ARE | 1 | **0.152±0.025** | 0.599± 0.286 | 1.000± 0.000 | 0.328± 0.033 |
| (↓) | 2 | **0.169±0.026** | 0.486± 0.121 | 1.000± 0.000 | 0.335± 0.028 |
| | 3 | **0.198±0.015** | 0.592± 0.162 | 1.000± 0.000 | 0.336± 0.029 |
| | 4 | **0.283±0.026** | 0.825± 0.060 | 1.000± 0.000 | 0.336± 0.032 |
| Fro. | 1 | **0.138±0.020** | 0.330± 0.167 | 0.509± 0.069 | 0.166± 0.035 |
| (↓) | 2 | **0.154±0.016** | 0.341± 0.078 | 0.695± 0.050 | 0.231± 0.031 |
| | 3 | **0.176±0.018** | 0.482± 0.106 | 0.817± 0.034 | 0.272± 0.032 |
| | 4 | **0.254±0.020** | 0.758± 0.075 | 0.902± 0.020 | 0.301± 0.034 |
| S.sim | 1 | **0.701±0.035** | 0.417± 0.216 | 0.269± 0.105 | 0.633± 0.110 |
| (↑) | 2 | **0.606±0.036** | 0.326± 0.097 | 0.106± 0.032 | 0.480± 0.072 |
| | 3 | **0.516±0.026** | 0.205± 0.074 | 0.055± 0.017 | 0.418± 0.043 |
| | 4 | 0.385± 0.048 | 0.072± 0.025 | 0.030± 0.006 | **0.401±0.021** |

### 4.3. *Experiments in multi-view kernel matrix completion*

We now describe our experiments on multi-view kernel matrix completion with unsupervised setting; i.e. there are no labels available and we assess the performance of the compared methods only on the matrix completion error measures introduced in the previous section. Thus, the MIC method is not applicable for comparison in this section.

#### 4.3.1. *Simulated data*

To validate our algorithm and to illustrate its generalization properties in predicting kernel values, we performed experiments with a simple simulated data set. We have created 100 data samples with a simple vector autoregression model of memory 1 where we periodically change the parameters of the model evolution, and constructed 7 views from overlapping column groups of the matrix to which the time series vectors have been stacked into. We calculated RBF kernels from these views. We consider a missing data scenario where every data sample is missing from randomly selected $a$ views, $a$ ranging from 1 to 4.

We report the results averaged over all the views for the various levels of missing data in Table 2, where we compare our CVKT to the other completion methods. To highlight the difference of our method to mean imputation that also performs relatively well with respect to the error measures, we show examples of completed kernel matrices in Fig. 2. Our method learns the overall trends in the kernel matrices, and is able to predict and generalize those.

#### 4.3.2. Drosophila melanogaster *pattern formation data set*

We now turn to a kernel completion task with a complex real-world multi-view dataset in order to validate our CVKT approach.

Image multiplexing is a relevant application of the cross-view kernel transfer method in biology. To study how cell fates are established by gene regulatory networks in the field of developmental biology, it has recently been proposed that a first necessary step is to integrate multiple views from heterogeneous image datasets [2]. Gene regulatory networks describe the sequence of interaction between various chemical species inside a cell or within a tissue, which ultimately lead to cell differentiation into a variety of functional types. The number of variables in these networks can go up to hundreds and each of them have to be measured separately with specific reporters. To understand the kinetics of these interactions it is necessary to reconstruct the time courses of their levels in various parts of the embryo. Despite many advances in microscopy techniques, it is still challenging to measure more than three of these variables at the same time, in addition, in the absence of reliable live reporters, some variables can only be measured in fixed images where the development is arrested, hence the need to integrate multiple views. As an illustration, live imaging of gastrulation provides information about nuclear positions as a function of time, but is silent about the levels of gene expression. On the other hand, an image of a fixed embryo reveals the distribution of an active enzyme but has no direct temporal information.

In the following example, we follow [2] and focus on the dorso-ventral patterning in *Drosophila melanogaster* early development. In this model system a graded profile of nuclear localization of a transcription factor named Dorsal (Dl) establishes the dorsoventral (DV) stripes of gene expression. Four datasets of fixed images were acquired to visualize nuclei (referred to as M, for morphology), protein expression of doubly phosphorylated ERK (dpERK, V1), Twist (V2), and Dorsal (V4), and mRNA expression of *ind* (V3) and *rho* (V5). The first dataset contains 108 images stained for dpERK and Twist. The second dataset contains 59 images stained for dpERK, *ind*, and Dorsal. The third dataset contains 58 images stained for dpERK, *ind*, and *rho*. The fourth dataset contains 30 images stained for Twist, *ind*, and *rho*. Examples of the images the data contains can be seen in Fig. 3. The distribution of the variables are shown on Fig. 4.[4]

In order to quantify the success of the proposed CVKT method, we select randomly samples to be missing for each of the views. The samples are selected in addition to the already missing samples, meaning that the selection is done in the teal coloured areas in Fig. 4. We then complete these samples with the information available in the other views. Note that we do not try to complete the truly missing samples, as our goal is to evaluate our algorithm and we want to be able to compare the completion results to known values. Thus for example when we consider view 2, we will only deal with datasets 1 and 4 (see Fig. 4), and we have five problems of different sizes. In addition to validating our method, this experiment mimics a real cross-validation situation when some samples in the data are truly missing.

[4] the view *ind* of the third dataset remains unused because of the lesser quality of the staining [2].

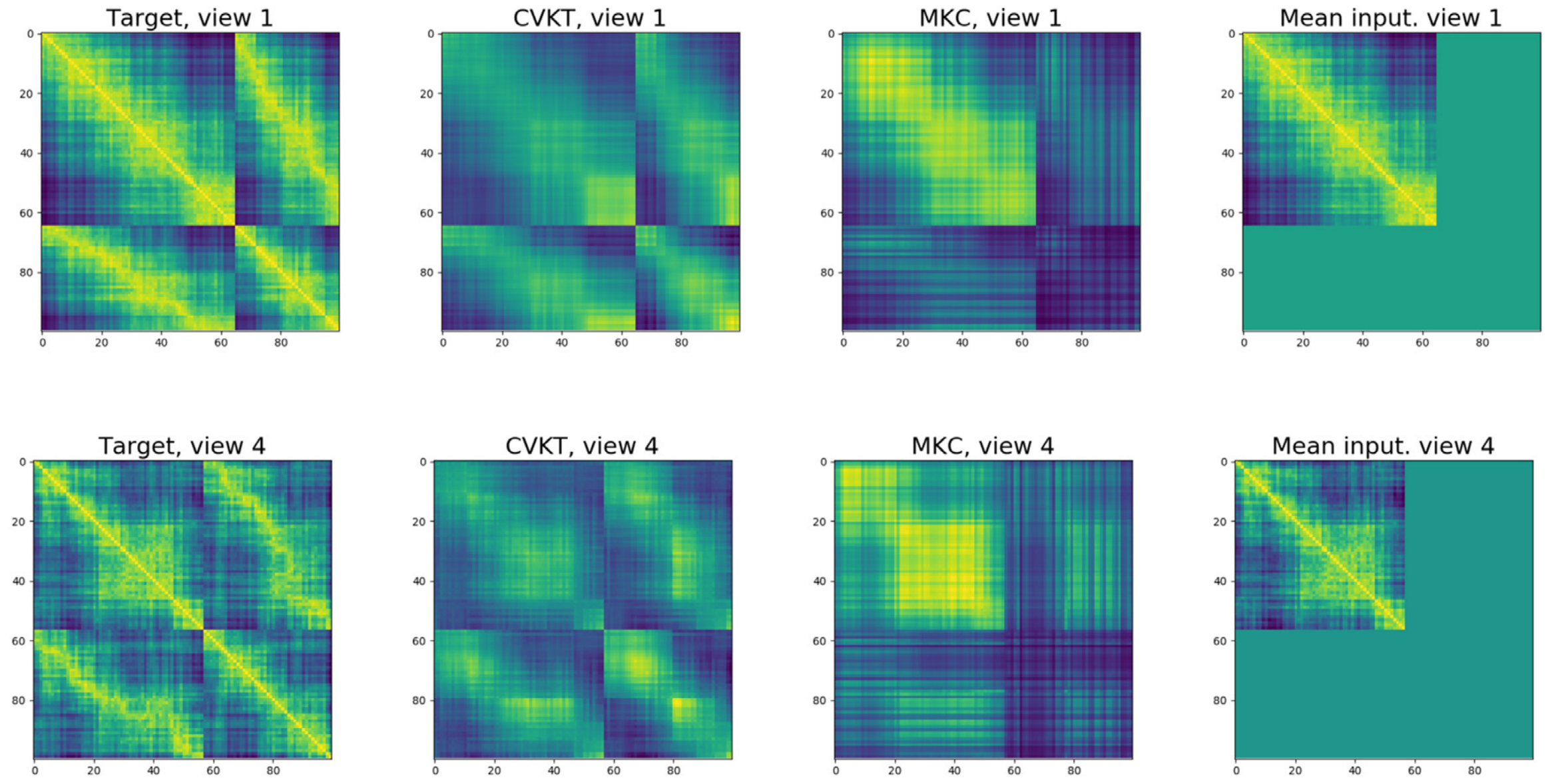

**Fig. 2.** Examples of target kernel matrices (left), our predicted kernel matrices (second from left), MKC completed kernel matrices (second from right) and mean imputed kernel matrices (right) on simulated data. On top row the matrices correspond to view 1 in the scenario when two views are missing per data sample, on the bottom row to view 4 in the scenario when three views are missing per data sample. The kernel matrices are reordered for better visualization such that top left corner contains the originally known data samples.

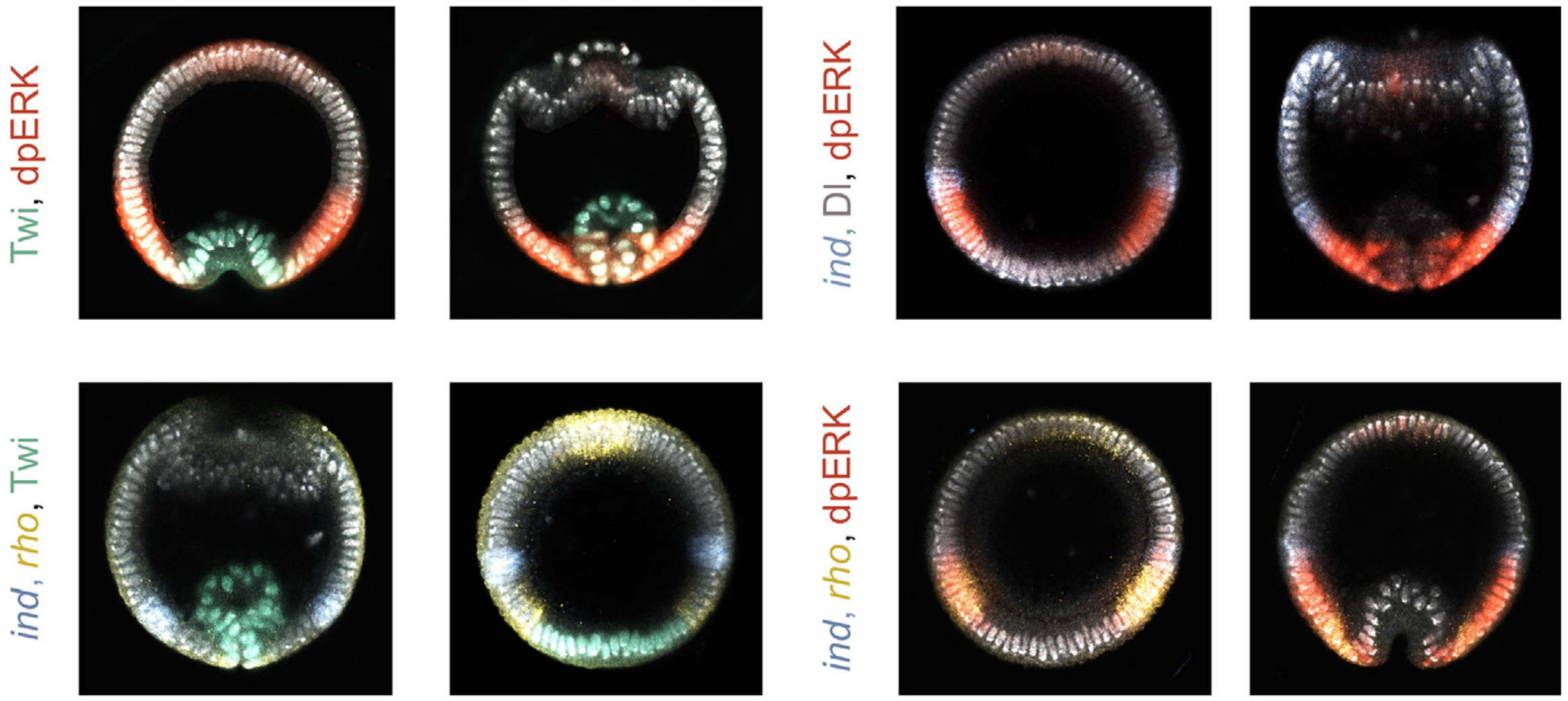

**Fig. 3.** Example images from the embryo dataset. In these images the colours identifying the views are modified so that they correspond over the datasets, e.g. dpERK is shown in red in all the images. In the dataset the views are highly correlated, a fact that can be exploited in the kernel completion task. Figure is adapted from Villoutreix et al. [2]. (For interpretation of the references to colour in this figure legend, the reader is referred to the web version of this article).

Our CVKT performs better in most of the views than other state-of-the-art methods with respect to CA error measure, shown in Table 3. Moreover, from Fig. 5 we can see that the structure of the kernel matrix is learned very well; however the exact values in our learned kernel matrices are slightly different ("lighter" images), which is no doubt then seen in the error measures. For the sake of clarity and brevity, we have focused in showing the case with 30% of missing data (a significant amount) in detail.

### 4.4. Classification accuracy with completed kernels

While it is good to analyse the performance of our method in only matrix completion task, it is important to remember that the reason for filling in the data in kernels is to make it possible to perform classification (or some other learning task) with them. Thus, in our next experiments our goal is to validate our CVKT as a kernel completion method also by applying the completed kernel matrices to their accompanying classification problem. This is done in order to highlight the differences between CVKT and mean imputation, methods producing very different results but for which the kernel completion error measures are sometimes very similar. We highlight that CVKT acts here as a preprocessing method for classification, as it only fills in the missing values in the multi-view kernel matrices. After applying CVKT (or other imputation method) we train a standard SVM classifier using the learned kernel matrices.

We consider the multiple features digits dataset[5] consisting of six views, as well as the uWaveGesture dataset[6] [37] containing three views. For the digits dataset, we selected 20 samples from all the 10 classes, resulting in six $200 \times 200$-sized kernel matrices for the completion problem. The views are various descriptions extracted from digit images, such as Fourier coefficients (view 'fou') or Karhunen-Loşve coefficients (view 'kar'). We use RBF kernels for views with data samples in $\mathbb{R}^d$, and Chi$^2$ kernels for views with data samples in $\mathbb{Z}^d$. The view 'mor' seems to[7] contain features fitting to both categorical and real data, so we consider a sum of the two appropriate kernels.

[5] https://archive.ics.uci.edu/ml/datasets/Multiple+Features

[6] http://www.timeseriesclassification.com/description.php?Dataset=UWaveGestureLibraryAll

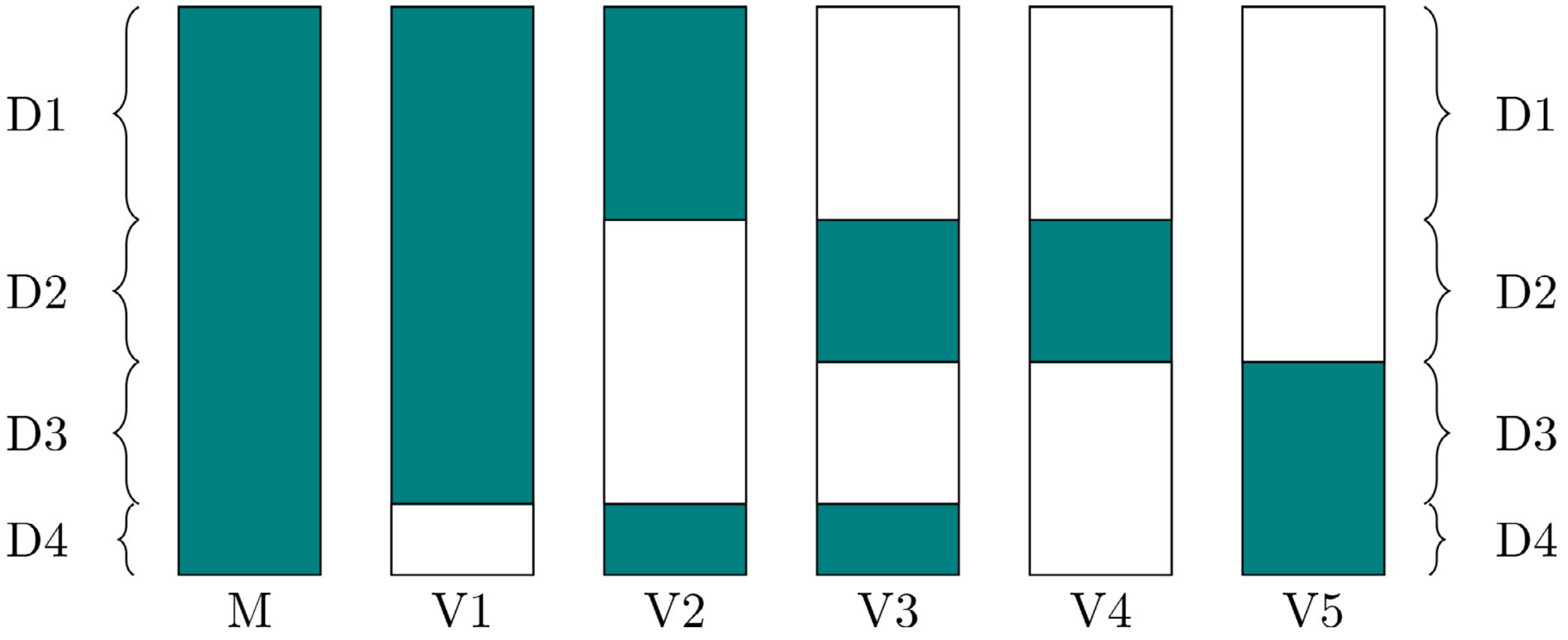

**Fig. 4.** Data availability in the views of *Drosophila melanogaster* data, coloured part referring to available data and white to missing; D to dataset and V to view. The datasets are of different sizes: 108, 59, 58 and 30 samples, respectively.

**Table 3**
Kernel matrix completion results on embryo data set [2] where 30% of available data is selected to be missing randomly per view. The arrow below the error measure shows whether higher values (↑) or lower values (↓) of the error measure indicate superior performance when comparing the various methods.

| Error measure | view | CVKT | MKC | zero-input. | mean-input. |
|---|---|---|---|---|---|
| CA | 1 | 0.295 | 0.230 | 0.254 | **0.206** |
| (↓) | 2 | 0.179 | 0.190 | 0.251 | **0.165** |
| | 3 | **0.162** | 0.244 | 0.259 | 0.166 |
| | 4 | **0.129** | 0.148 | 0.246 | 0.151 |
| | 5 | **0.132** | 0.170 | 0.225 | 0.164 |
| ARE | 1 | **0.843** | 0.966 | 1.000 | 0.919 |
| (↓) | 2 | **0.723** | 0.915 | 1.000 | 0.842 |
| | 3 | **0.739** | 0.968 | 1.000 | 0.831 |
| | 4 | **0.690** | 0.805 | 1.000 | 0.820 |
| | 5 | **0.734** | 0.884 | 1.000 | 0.882 |
| Fro | 1 | 0.717 | 0.647 | 0.666 | **0.608** |
| (↓) | 2 | 0.574 | 0.608 | 0.663 | **0.551** |
| | 3 | **0.546** | 0.656 | 0.671 | 0.551 |
| | 4 | **0.490** | 0.528 | 0.657 | 0.529 |
| | 5 | **0.497** | 0.559 | 0.632 | 0.549 |
| S.sim | 1 | 0.526 | 0.584 | **0.692** | 0.672 |
| (↑) | 2 | **0.740** | 0.641 | 0.554 | 0.704 |
| | 3 | **0.722** | 0.571 | 0.519 | 0.673 |
| | 4 | **0.730** | 0.703 | 0.530 | 0.690 |
| | 5 | **0.636** | 0.566 | 0.602 | 0.566 |

We randomly set samples to be assumed missing in this dataset. We vary the level of total missing samples in the whole dataset from 10% to 50%, by taking care that all the samples are observed at least in one view, and that all views have observed samples. We note that in order to fill in missing values for a given view $v$, we need data at least from one other view to learn the transformation from. Thus we cannot consider arbitrarily high levels of missing data in our experiments. For example with 3-view data this threshold would be 33% missing values; with 5-view data 60% missing values. When we consider levels higher than this threshold not all the samples will end up having enough data to be used in learning the transformation (i.e. they will only be observed in one view), and in essence our training set size diminishes. For example the uWaveGesture experiments end up operating in this regime for the highest levels of missing data.

After we perform kernel completion with CVKT and the competing methods, we give the completed matrices (selected again w.r.t. highest CA) to SVM classifiers. For CVKT the selection based on CA was done individually for all the views, since it performs individual optimization. For MKC the errors were averaged over the views, and the result with lowest overall error was chosen, as MKC performs joint optimization. The MIC method, originally introduced for incomplete multi-view clustering, builds a common representation of the views that can be then used in the classification task, while the individual views rest incomplete. Thus, with this method we cannot compare the view-specific performance, but show comparisons to this common representation. With MIC we also use SVM classifier, with RBF kernel. In order to perform classification we divide the data in half for training and testing, and this selection is the same for all the kernel matrices. Both training and testing sets contain samples for which the views were assumed missing in completion task.

We report the accuracies on test data averaged over five different selections for missing data in Fig. 6 for the digits dataset. Our CVKT performs the classification superiorly to other kernel completion methods, and comparably to using the original fully known kernel matrix up to the case with 30% of missing data. The MIC method does not perform as well as CVKT in most of the views, except in two where it is the same with the lowest amount of data missing. However even in these views its performance drops much more rapidly with level of missing data than for example CVKT, and for 50% of missing data it performs always worse than even mean or zero inputed kernels.

In previous experiments the mean imputation has sometimes performed similarly to CVKT with respect to matrix completion error measures. It is the case also with the digits dataset (see Table 4), but the classification accuracy CVKT obtains is consistently higher than that of mean imputation (see Fig. 6). This is as expected; the inputed mean values do not carry meaningful information about the data samples they are supposed to represent, and thus will not allow for successful classification. It is interesting to notice that for view 'fou', the classification accuracy after completing 10% missing data is higher with CVKT kernel than with the original full kernel matrix. It might be that in this case CVKT has been able to filter out some noise distortions in samples, which could give it better performance than the baseline. This could be analogous to using kernel approximation schemes as regularization [38]. We emphasize that in the experiments the kernel matrix completion is done fully independently from the consecutive classification task, without knowing which samples would be used in training and which in testing.

[7] According to the data source, the source image dataset is lost, and there is very little information on the views.

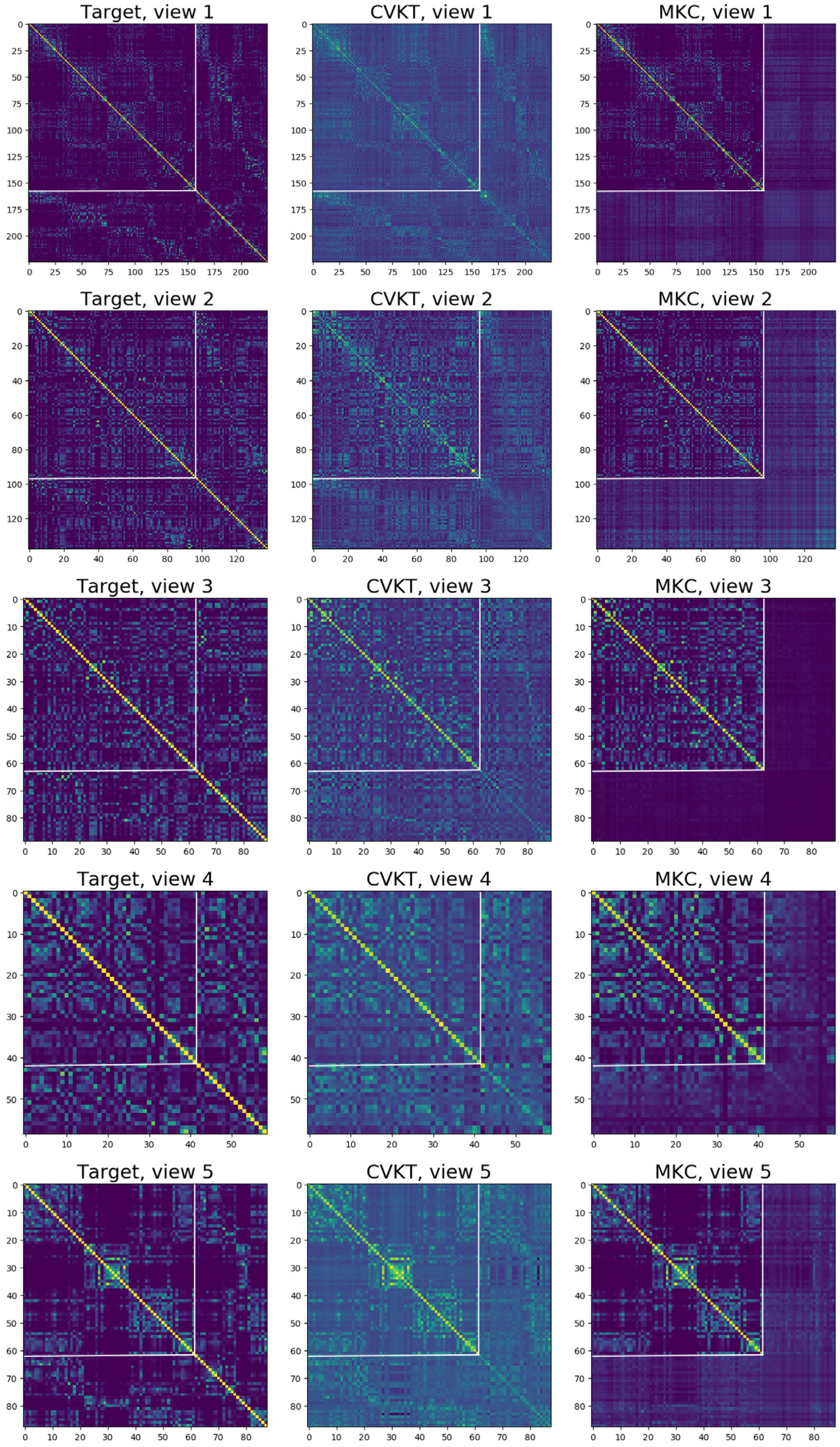

**Fig. 5.** Target kernel matrices (left), our predicted kernel matrices (middle) with CVKT, and MKC predicted kernel matrices (right) of embryo data [2] when randomly selected 30% of the available samples were set to be missing. The kernel matrices are reordered for better visualization such that top left corner contains the originally known data samples (areas with unknown and known samples are separated with white lines).

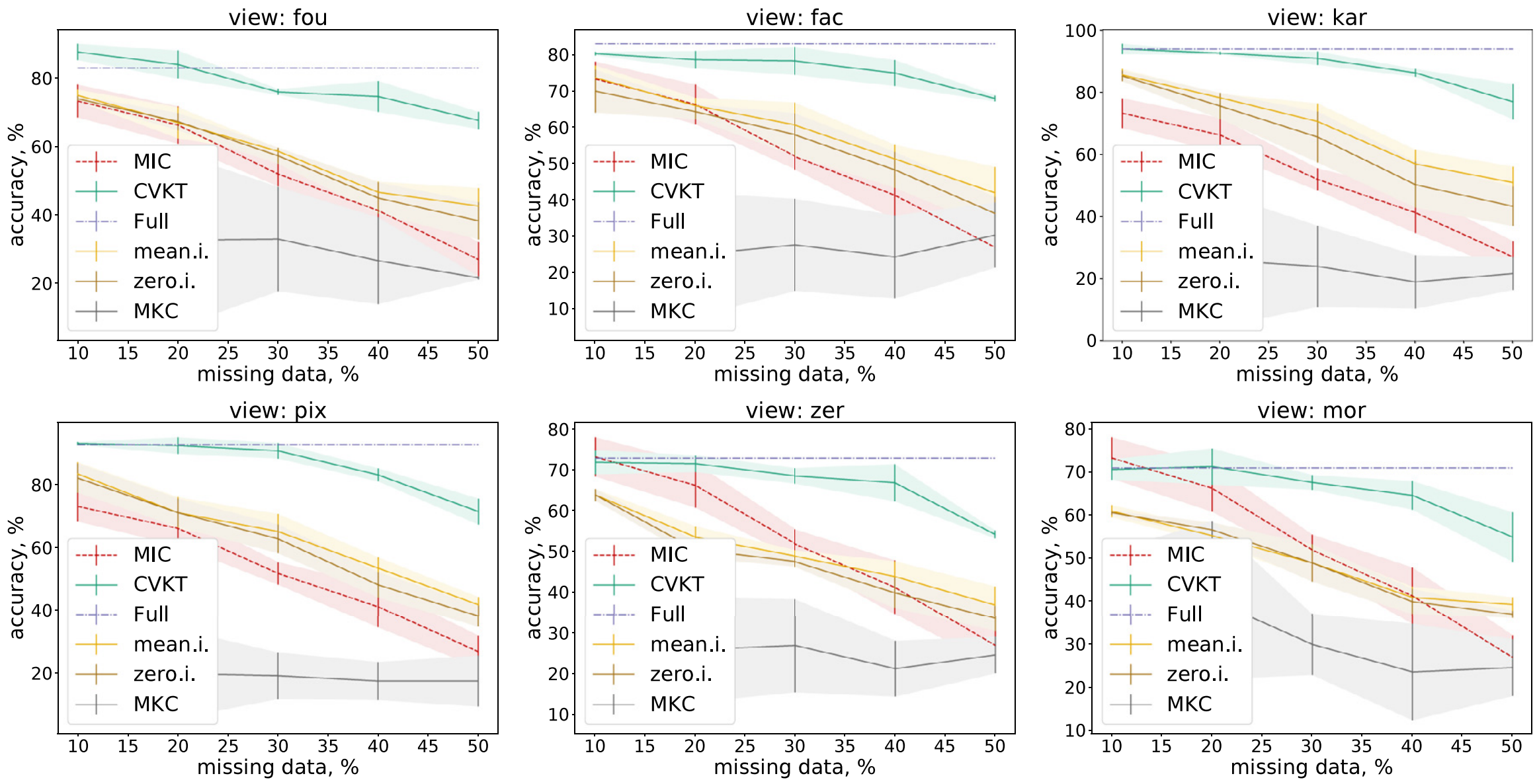

**Fig. 6.** Accuracies of classification with full, mean inputed, zero inputed, CVKT-completed and MKC-completed kernel matrices for all six views of the digits dataset as a function of level of missing data in views. The MIC results are for a common representation, and thus identical in all the plots.

**Table 4**
Completion error measures on digits data set with various levels of missing data samples in the views, averaged over the views. The arrow below the error measure shows whether higher values (↑) or lower values (↓) of the error measure indicate superior performance when comparing the various methods.

| Error measure | missing % | CVKT | MKC | zero-input. | mean-input. |
|---|---|---|---|---|---|
| CA | 10 | 0.0097±0.0059 | 0.214±0.281 | 0.097±0.021 | **0.004±0.002** |
| (↓) | 20 | 0.0104±0.0062 | 0.147±0.205 | 0.195±0.026 | **0.008±0.004** |
| | 30 | **0.012±0.006** | 0.143±0.043 | 0.295±0.031 | **0.012±0.005** |
| | 40 | **0.014±0.007** | 0.189±0.051 | 0.392±0.035 | **0.015±0.007** |
| | 50 | **0.018±0.009** | 0.232±0.064 | 0.493±0.042 | **0.017±0.008** |
| ARE | 10 | **0.148±0.054** | 4.916±12.008 | 1.000±0.000 | 0.217±0.057 |
| (↓) | 20 | **0.155±0.049** | 1.808±4.356 | 1.000±0.000 | 0.213±0.052 |
| | 30 | **0.167±0.048** | 0.739±0.112 | 1.000±0.000 | 0.214±0.052 |
| | 40 | **0.181±0.048** | 0.790±0.141 | 1.000±0.000 | 0.214±0.053 |
| | 50 | **0.197±0.052** | 0.851±0.269 | 1.000±0.000 | 0.213±0.052 |
| Fro | 10 | 0.131±0.044 | 2.579±6.389 | 0.427±0.045 | **0.090±0.024** |
| (↓) | 20 | 0.137±0.045 | 1.914±5.539 | 0.591±0.036 | **0.124±0.030** |
| | 30 | **0.146±0.042** | 0.545±0.092 | 0.708±0.031 | **0.148±0.035** |
| | 40 | **0.163±0.044** | 0.631±0.121 | 0.793±0.027 | 0.167±0.039 |
| | 50 | **0.182±0.047** | 0.731±0.246 | 0.860±0.026 | **0.181±0.042** |
| S.sim | 10 | **0.760±0.117** | 0.304±0.150 | 0.393±0.088 | **0.862±0.050** |
| (↑) | 20 | 0.730±0.114 | 0.252±0.132 | 0.185±0.057 | **0.755±0.068** |
| | 30 | **0.678±0.115** | 0.146±0.072 | 0.105±0.039 | 0.660±0.087 |
| | 40 | **0.598±0.124** | 0.101±0.047 | 0.069±0.025 | 0.581±0.092 |
| | 50 | **0.509±0.134** | 0.078±0.027 | 0.048±0.016 | **0.509±0.104** |

We follow the same experimental protocol also for the uWaveGesture dataset, where we consider the 896 training samples in three views, with 8 classes. We report the completion error measures in Table 5. Again, according to some error measures, the mean imputation seems to be performing better than the dedicated matrix completion methods. However, again, from Fig. 7, we can see that CVKT performs better in the subsequent classification task in most of the cases. It is clear that CVKT retains more relevant information about the data than the simple imputation methods, yet this is not always reflected in the completion error measures. The MIC method performs here better than other baselines, but were we to consider MKL-style combinations of the individual kernels used in experiments of Fig. 7, the performance of CVKT and MIC would be almost identical.

Furthermore, we performed statistical testing to assess the significance of our classification results, excluding the MIC method since it cannot be used for individual views, nor can it be used in the task of kernel matrix completion. First of all, we consider the McNemar test. We compared the CVKT-based classification to the four other methods: classification with full kernels, MKC-filled kernels, and mean and zero-inputed kernels. We show in Tables 6 and 7 the obtained p-values, and also how often the null hypothesis was rejected (p-value threshold 0.05), i.e. how often the two classification results were significantly different. We observe that the differences mostly grow with the level of missing values. For the digits dataset, the CVKT results with 10% and 20% of missing data are almost indistinguishable from the full classification according to the test, and the mean and zero imputation results are very different from those obtained with CVKT.

**Table 5**
Completion error measures on uWaveGesture data set with various levels of missing data samples in the views, averaged over the views. The arrow below the error measure shows whether higher values (↑) or lower values (↓) of the error measure indicate superior performance when comparing the various methods.

| Error measure | missing % | CVKT | MKC | zero-input. | mean-input. |
|---|---|---|---|---|---|
| CA | 10 | 0.021±0.001 | 0.101±0.031 | 0.097±0.009 | **0.006±0.001** |
| (↓) | 20 | 0.023±0.001 | 0.099±0.037 | 0.192±0.012 | **0.011±0.001** |
| | 30 | 0.025±0.002 | 0.184±0.100 | 0.302±0.022 | **0.016±0.003** |
| | 40 | 0.027±0.003 | 0.223±0.106 | 0.379±0.041 | **0.019±0.003** |
| | 50 | 0.028±0.003 | 0.252±0.078 | 0.452±0.072 | **0.022±0.001** |
| ARE | 10 | **0.198±0.004** | 0.889±0.146 | 1.000±0.000 | 0.246±0.018 |
| (↓) | 20 | **0.207±0.005** | 0.679±0.178 | 1.000±0.000 | 0.245±0.017 |
| | 30 | **0.217±0.009** | 0.750±0.180 | 1.000±0.000 | 0.244±0.015 |
| | 40 | **0.226±0.012** | 0.770±0.171 | 1.000±0.000 | 0.243±0.015 |
| | 50 | **0.229±0.011** | 0.757±0.175 | 1.000±0.000 | 0.244±0.015 |
| Fro | 10 | 0.205±0.004 | 0.587±0.204 | 0.430±0.019 | **0.108±0.009** |
| | 20 | 0.213±0.005 | 0.468±0.086 | 0.589±0.016 | **0.147±0.009** |
| | 30 | 0.222±0.009 | 0.559±0.128 | 0.716±0.022 | **0.178±0.014** |
| | 40 | 0.230±0.012 | 0.615±0.118 | 0.782±0.034 | **0.194±0.013** |
| | 50 | 0.235±0.011 | 0.664±0.090 | 0.832±0.050 | **0.206±0.003** |
| S.sim | 10 | 0.466±0.064 | 0.244±0.176 | 0.508±0.033 | **0.898±0.010** |
| (↑) | 20 | 0.424±0.057 | 0.192±0.143 | 0.294±0.037 | **0.793±0.014** |
| | 30 | 0.362±0.047 | 0.217±0.059 | 0.178±0.014 | **0.670±0.025** |
| | 40 | 0.304±0.021 | 0.149±0.061 | 0.129±0.029 | **0.588±0.049** |
| | 50 | 0.278±0.026 | 0.108±0.034 | 0.099±0.044 | **0.503±0.075** |

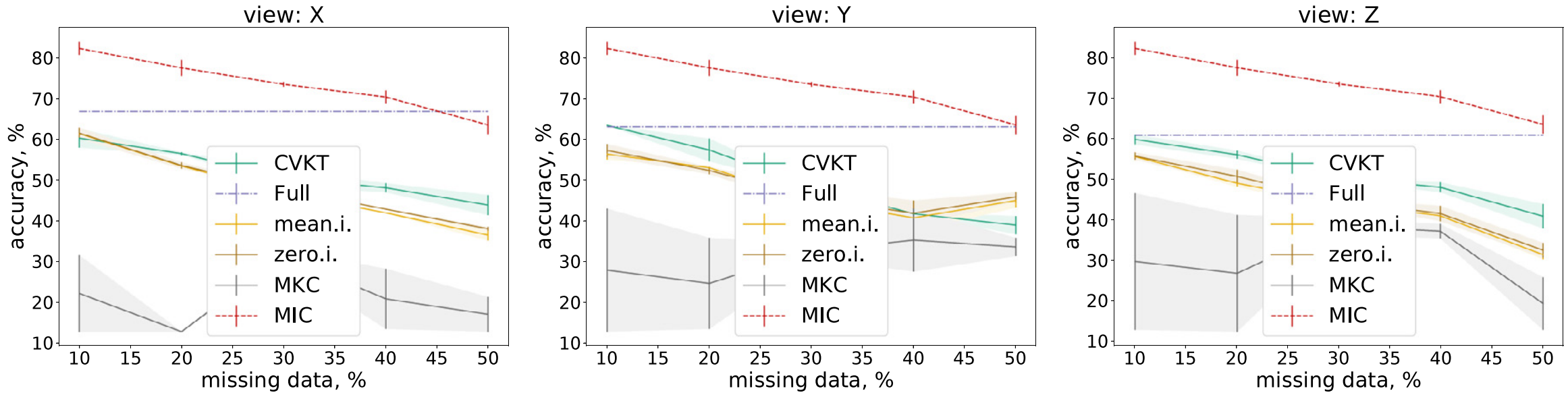

**Fig. 7.** Accuracies of classification with full, mean inputed, zero inputed, CVKT-completed and MKC-completed kernel matrices for all three views of the uWaveGesture dataset as a function of level of missing data in views. The MIC results are for a common representation, and thus identical in all the plots.

**Table 6**
McNemar's test on various classification results compared to CVKT classification results with the digits dataset. The table displays the average p-values ± its standard deviation, and in parenthesis as percentage with how many of the runs McNemar's test rejects the null hypothesis (i.e. the results can be said to be statistically significantly different) with p-value threshold at 0.05.

| missing % | full | MKC | zero-input. | mean-input. |
|---|---|---|---|---|
| all | 0.299 ± 0.317 (33.3) | 0.000 ± 0.001 (100.0) | 0.023 ± 0.086 (90.7) | 0.023 ± 0.086 (90.7) |
| 10% | 0.505 ± 0.284 (3.3) | 0.000 ± 0.001 (100.0) | 0.066 ± 0.104 (63.3) | 0.066 ± 0.104 (63.3) |
| 20% | 0.445 ± 0.310 (3.3) | 0.000 ± 0.002 (100.0) | 0.031 ± 0.148 (96.7) | 0.031 ± 0.148 (96.7) |
| 30% | 0.362 ± 0.299 (20.0) | 0.000 ± 0.000 (100.0) | 0.009 ± 0.039 (96.7) | 0.009 ± 0.039 (96.7) |
| 40% | 0.171 ± 0.265 (50.0) | 0.000 ± 0.000 (100.0) | 0.002 ± 0.006 (100.0) | 0.002 ± 0.006 (100.0) |
| 50% | 0.011 ± 0.025 (90.0) | 0.000 ± 0.000 (100.0) | 0.004 ± 0.014 (96.7) | 0.004 ± 0.014 (96.7) |

**Table 7**
McNemar's test on various classification results compared to CVKT classification results with the uWaveGesture dataset. The table displays the average p-values ± its standard deviation, and in parenthesis as percentage with how many of the runs McNemar's test rejects the null hypothesis (i.e. the results can be said to be statistically significantly different) with p-value threshold at 0.05.

| missing % | full | MKC | zero-input. | mean-input. |
|---|---|---|---|---|
| all | 0.071 ± 0.200 (83.3) | 0.007 ± 0.034 (96.7) | 0.272 ± 0.317 (33.3) | 0.272 ± 0.317 (33.3) |
| 10% | 0.337 ± 0.334 (16.7) | 0.000 ± 0.000 (100.0) | 0.387 ± 0.327 (33.3) | 0.387 ± 0.327 (33.3) |
| 20% | 0.016 ± 0.011 (100.0) | 0.000 ± 0.001 (100.0) | 0.166 ± 0.213 (50.0) | 0.166 ± 0.213 (50.0) |
| 30% | 0.000 ± 0.000 (100.0) | 0.000 ± 0.000 (100.0) | 0.376 ± 0.379 (16.7) | 0.376 ± 0.379 (16.7) |
| 40% | 0.000 ± 0.000 (100.0) | 0.004 ± 0.009 (100.0) | 0.286 ± 0.309 (16.7) | 0.286 ± 0.309 (16.7) |
| 50% | 0.000 ± 0.000 (100.0) | 0.032 ± 0.071 (83.3) | 0.146 ± 0.245 (50.0) | 0.146 ± 0.245 (50.0) |

**Table 8**
Summary of the results of Friedman-Nemenyi test, showing if CVKT results are statistically significantly different (in bold: test value larger than critical difference, "CD") to other compared multi-view kernel completion methods (mean and zero imputation, MKC) with uWaveGesture and digits datasets, with $\alpha = 0.1$ and $\alpha = 0.05$.

| | error measure | CD | MKC | mean i. | zero.i. |
|---|---|---|---|---|---|
| $\alpha = 0.1$ | CA | 1.32 | **1.35** | 0.7 | **1.95** |
| | ARE | 1.32 | **2.2** | 1.0 | **2.8** |
| | Accuracy | 0.62 | **2.91** | **1.28** | **1.46** |
| $\alpha = 0.05$ | CA | 1.48 | 1.35 | 0.7 | **1.95** |
| | ARE | 1.48 | **2.2** | 1.0 | **2.8** |
| | Accuracy | 0.70 | **2.91** | **1.28** | **1.46** |

Secondly, we perform the Friedman-Nemenyi test [39] on uWave gesture and digits datasets in both classification and kernel matrix completion settings, in order to verify if the results of the different methods are overall statistically significantly different. Here we consider the various levels of missing data as different datasets from the point of view of the test, in classification we also consider the different views. In kernel completion we consider only the CA and ARE error measures, as the Frobenius norm error and structural similarity index give very similar results to the CA measure.

First, for all the error measures, we perform the Friedman test. For this we compute the Friedman statistic

$$\mathcal{X}_F^2 = \frac{12N}{k(k+1)}\left[\sum_j [R_x]_j^2 - \frac{k(k+1)^2}{4}\right],$$

in which $R_x$ is either $R_{CA}$, $R_{ARE}$ or $R_{acc}$, and $[R_x]_j$ stands for average ranking for the method $j$; the mean values of the rankings with CA, ARE or accuracy measure. As we consider four algorithms, $k = 4$, and $N$ stands for the number of experiments. For CA and ARE as we have averaged the results over the views $N = 10$ (both datasets are considered with five different levels of missing data), while with accuracy score we perform and show classification with the views independently: $N = 45$. We can use directly the Friedman statistic in rejecting the null hypothesis, or as it is somewhat conservative (see [39] and references therein), we can consider

$$F_F = \frac{(N-1)\chi_F^2}{N(k-1) - \chi_F^2},$$

and reject the null hypothesis by comparing its values to the critical values of f-distribution.

After observing that the null hypothesis is in all the cases rejected, we proceed with the pairwise comparisons (CVKT to MKC, mean imputation and zero imputation) and perform the Nemenyi test by comparing the differences of the average rankings to the critical difference value

$$CD = q_\alpha\sqrt{\frac{k(k+1)}{6N}}$$

with both $\alpha = 0.05$ and $\alpha = 0.1$. From this, we obtain information whether the results of the two algorithms are statistically significantly different or not. We summarize the results in Table 8. It is easy to conclude that while the matrix completion error measures have not necessarily shown much difference between CVKT and mean imputation (or MKC with $\alpha = 0.05$), the performance difference measured in classification accuracy clearly shows the superior performance of CVKT.

## 5. Conclusion

We have introduced a novel idea for performing multi-view kernel matrix completion by transferring cross-view knowledge to represent the views with missing values. We learn to represent the kernels with features of other views linearly transformed to a new feature space. This allows predicting the missing values of a kernel with features available in the other views. Our algorithm solves the problem efficiently, since the views can be treated individually, and no heavy joint optimization is performed. This individual treatment of views also gives more flexibility to our approach. As our experiments with simulated and real data demonstrate, our method is able to find generalizable structures from the incomplete kernel matrices, and is able to predict those structures in completing them. Our method completes the kernel matrices in a way that allows using them successfully in machine learning applications, as demonstrated with experiments on datasets of handwritten digits and images of flowers. The competing method, MKC, performed worse than expected. It might be that the assumptions of the chosen algorithm, $\text{MKC}_{embd(ht)}$, are not optimal for this specific problem, and one of the slower ones would have performed better. In [22] it is assumed that each view has a small basis set of samples with which the view can be characterized, and it might not be the case in our experiments. Additionally, the experimental setting is challenging with a lot of missing data samples. As the data is randomly missing from views for some data samples, even in lower levels of missing data, only one or two views might be available.

Our experiments propose that the current metrics to evaluate the matrix completion results are not fully usable by themselves. Two very different approaches can give similar errors on kernel completion, but give widely different accuracies on application to classification. One possible line of future work would be studying how one could better quantify the success of the kernel completion task.

As a successful multi-view kernel completion method, this work opens up novel avenues of research also for the reconstruction of the initial data samples. As multi-view kernel learning method, it would be interesting to further study the suitability of feature transfer, for example in aligning the features with ideal kernel formed on the labels. This might prove a competitive way to form a multi-view kernel, compared to the currently widely used multiple kernel learning framework. Also, investigating the connections to operator-valued kernels on multi-view setting with missing data could be a possible way to move forward with this research.

## Declaration of Competing Interest

None.

## Acknowledgements

This work is mainly granted by the french national project ANR Lives ANR-15-CE23-0026, and by the Turing Center for Living Systems (CENTURI) for PV. For the most part work by RH has been done in Aix-Marseille University – the part in Aalto University has been funded by Academy of Finland grants 334790 (MAGITICS) and 310107 (MACOME).